\documentclass[11pt]{article}
\usepackage{acl2020}
\usepackage{times}
\usepackage{url}
\usepackage{latexsym}
\usepackage{capt-of}% or \usepackage{caption}
\usepackage{booktabs}
\usepackage{varwidth}
\newsavebox\tmpbox
\usepackage[inline]{enumitem}
\usepackage{floatrow}
\newfloatcommand{capbtabbox}{table}[][\FBwidth]

\usepackage{subfig}
\usepackage{microtype}

\aclfinalcopy % Uncomment this line for the final submission
\usepackage{caption}

\usepackage{graphicx}
\usepackage{tabularx}
\usepackage{soul}
\usepackage{color}
\usepackage{epstopdf}
\usepackage[latin1]{inputenc}

\usepackage{hyperref}
\usepackage{xstring}
\usepackage{comment}
\usepackage{flushend}

\usepackage[inline]{enumitem}

\newcommand{\secref}[1]{\StrSubstitute{\getrefnumber{#1}}{.}{ }}

\title{Enriching the Transformer with Linguistic Factors\\for Low-Resource Machine Translation} %\vspace*{.5\baselineskip} \normalfont{ The Title \ul{Must Be} Capitalised as in:\\ \vspace*{.5\baselineskip} \textbf{The Rise and Fall of Ziggy Stardust and the Spiders from Mars}}}

\author{Jordi Armengol-Estap\'e, Marta R. Costa-juss\`a, Carlos Escolano\\
TALP Research Center, Universitat Polit\`ecnica de Catalunya, Barcelona \\
         \tt{jordi.armengol.estape@gmail.com,\{marta.ruiz,carlos.escolano\}@upc.edu}}

\begin{document}

\maketitle
\begin{abstract}
    Introducing factors, that is to say, word features such as linguistic information referring to the source tokens, is known to improve the results of neural machine translation systems in certain settings, typically in recurrent architectures. This study proposes enhancing the current state-of-the-art neural machine translation architecture, the Transformer, so that it allows to introduce external knowledge. In particular, our proposed modification, the Factored Transformer, uses linguistic factors that insert additional knowledge into the machine translation system. Apart from using different kinds of features, we study the effect of different architectural configurations. Specifically, we analyze the performance of combining words and features at the embedding level or at the encoder level, and we experiment with two different combination strategies. With the best-found configuration, we show improvements of 0.8 BLEU over the baseline Transformer in the IWSLT German-to-English task. Moreover, we experiment with the more challenging FLoRes English-to-Nepali benchmark, which includes both extremely low-resourced and very distant languages, and obtain an improvement of 1.2 BLEU.% These improvements are achieved with linguistic and not with semantic information. %(which accounts for an increase of almost 40\%).  

\end{abstract}

\section{Introduction}
Many classical Natural Language Processing (NLP) pipelines used either linguistic features \cite{koehn-hoang-2007-factored,du-etal-2016-using}. In recent years, the rise of neural architectures has diminished the importance of the aforementioned features. Nevertheless, some works have still shown the effectiveness of introducing linguistic information into neural machine translation systems, typically in recurrent sequence-to-sequence (Seq2seq) architectures \cite{DBLP:journals/corr/SennrichH16,DBLP:journals/corr/Garcia-Martinez16,espanaVanGenabith:LREC:2018}.
By factored Neural Machine Translation (NMT), we refer to the use of word features alongside the words themselves to improve translation quality. Both the encoder and the decoder of a Seq2seq architecture can be modified to obtain better translations \cite{DBLP:journals/corr/Garcia-Martinez16}. The most prominent approach consists of modifying the encoder such that instead of only one embedding layer, the encoder has as many embedding layers as factors, one for words themselves and one for each feature, and then the embedding vectors are concatenated and input to the rest of the model, which remains unchanged \cite{DBLP:journals/corr/SennrichH16}. The embedding sizes are set according to the respective vocabularies of the features. %This approach consisted in modifying the standard Seq2seq with the aim of inputting an arbitrary amount of aligned source sequences, one for the source words and the rest for each one of the features. In this embedding-level modification, each feature had its own embedding layer, and the embedding size was set according to the vocabulary. %(ie. features with tiny vocabulary sizes had very low-dimensional embeddings). 
%The output of the different embedding layers was concatenated and input to the encoder.
Notice that Byte Pair Encoding (BPE) \cite{sennrich-etal-2016-neural}, an unsupervised preprocessing step for automatically splitting words into subwords with the goal of improving the translation of rare or unseen words, was applied to the words here. Thus, the features had to be repeated for each subword. %Otherwise, features would be misaligned. An additional feature, subword tags, was introduced for denoting the position of the subword in the original word (eg. 'B' for the beginning of the word). 
In \cite{espanaVanGenabith:LREC:2018}, the exact same architecture was used, except that this new proposal used concepts extracted from a linked data database, BabelNet \cite{NavigliPonzetto:12aij}. These semantic features, synsets, were shown to improve zero-shot translations.
%A recent work proposed using concepts extracted from BabelNet \cite{NavigliPonzetto:12aij}, which is a linked data database with semantic relations, instead of classical linguistic features \cite{espanaVanGenabith:LREC:2018}. 
All the cited works obtained moderate improvements with respect to the BLEU scores of the corresponding baselines. %Some additional advantages were observed. In particular, in the case of \cite{espanaVanGenabith:LREC:2018}, the use of concepts from BabelNet improves the results of zero-shot translations. %\cite{NavigliPonzetto:12aij} improves the results of zero-shot translations.
%Factored NMT is known to obtain moderate improvements in translation quality. In \cite{DBLP:journals/corr/SennrichH16}, %However, up to this point and to the best of our knowledge, all the published works on this matter are based on Recurrent Neural Networks (RNNs) instead of the Transformer \cite{vaswani2017attention}, which is the current state-of-the-art architecture for NMT for both translation quality and efficiency. %Investigating whether the Transformer can take advantage of factors is indeed highly relevant. 
%Additionally, and unlike previous works, we are focusing our attention on low-resourced datasets, which is a big challenge for deep learning techniques in general, since they rely on large quantities of data.
Some works have previously proposed additional ways to combine sources and introduce hierarchical linguistic information in \cite{currey-heafield-2019-incorporating,currey-heafield-2018-multi,libovicky-etal-2018-input,tebbifakhr-etal-2018-multi}.
%Thank you for the references, we will add them in the camera-ready copy. We were not aware of them since we were focused on the topic of linguistic and semantic features for factored NMT. We acknowledge that the intersection between these works and ours is not empty, but there are some differences. For instance, in Multi-source Transformer with Combined Losses for Automatic Post-Editing, they studied 2 encoders for two sentences with concatenation, but we tried with sum as well and targeted linguistic features. In Incorporating Source Syntax into Transformer-Based 
The main goal of this work, and differently from previous works using NMT architectures based on recurrent neural networks, is to modify the Transformer to make it compatible with factored NMT with an architecture that we call Factored Transformer and inject linguistic knowledge %from lemmas, which is the best performing feature in \cite{DBLP:journals/corr/SennrichH16}, 
and concepts extracted from the semantic linked data database, BabelNet \cite{NavigliPonzetto:12aij}. We are focusing our attention on low-resource datasets. %, which is a big challenge for deep learning techniques in general, since they rely on large quantities of data.%The use of classical linguistic features was planned as well. %We were expecting that the use of the Factored Transformer with certain features would improve the baseline, which was based on the vanilla Transformer. %To be more specific, the particular contributions of this work are:

%On the other hand, concepts from BabelNet have only been used in NMT in one recent study \cite{espanaVanGenabith:LREC:2018}. Since this semantic information improved zero-shot translations, we believe that it can be particularly useful in low-resource settings. Deep learning has the drawback of requiring large quantities of data, while parallel corpora are expensive, and many languages do not have many resources for NLP. Previous works in factored NMT did not particularly focus in low-resource settings, unlike this work.

\section{Factored Transformer}
\label{obj}

%\subsection{Designing and implementing a new architecture for NMT, the Factored Trasformer}  
Unlike the vanilla Transformer \cite{vaswani2017attention}, the Factored Transformer can work with factors; that is, instead of just being input the original source sequence, it can work with an arbitrary number of feature sequences. Those features can be injected at embedding-level, as in the previous works we described above (but in a Transformer instead of a recurrent-based seq2seq architecture), or at the encoder level. We have implemented the two model variants.

\textbf{1-encoder model (depicted in Figure \ref{figure1}, left):} Each factor, including the words themselves, has its own embedding layer. The embedding vectors of the different factors are combined, positional encoding is summed and input to the following layer. The rest of the model remains unchanged. The positional encoding is summed to the combined vector and not to each individual embedding because we are not modifying the length of the sequence; therefore, the relative positions remain unchanged.

\textbf{N-encoders model (depicted in Figure \ref{figure1}, right):} We intuited that features with large vocabulary sizes %, like BabelNet concepts \cite{NavigliPonzetto:12aij} (see section \ref{sec:features}), have large vocabulary sizes, 
could benefit from having a specific encoder. In this variant, each factor has its own full encoder (instead of just its own embedding layer). The outputs from the encoder are combined and input to the following layer. The rest of the model remains unchanged.

Once we have the outputs of the multiple embedding layers (the 1-encoder) or the N-encoders, they must be aggregated before being input to the next layer. We have considered two combination strategies: 

\textbf{Concatenation:} The outputs of the different embedding layers or encoders are concatenated.

\textbf{Summation:} The outputs of the different embedding layers or encoders are summed.

In both cases, the dimensions must agree. The decoder embedding size must be equal to the encoder embedding size. If the outputs from the different encoders or embedding layers are concatenated, they do not need to have the same embedding size, but the resulting embedding size is increased. Instead, if they are summed, they must share the same dimensionality, but the resulting vector size is not increased.
\ffigbox{%
  \includegraphics[width=0.5\textwidth]{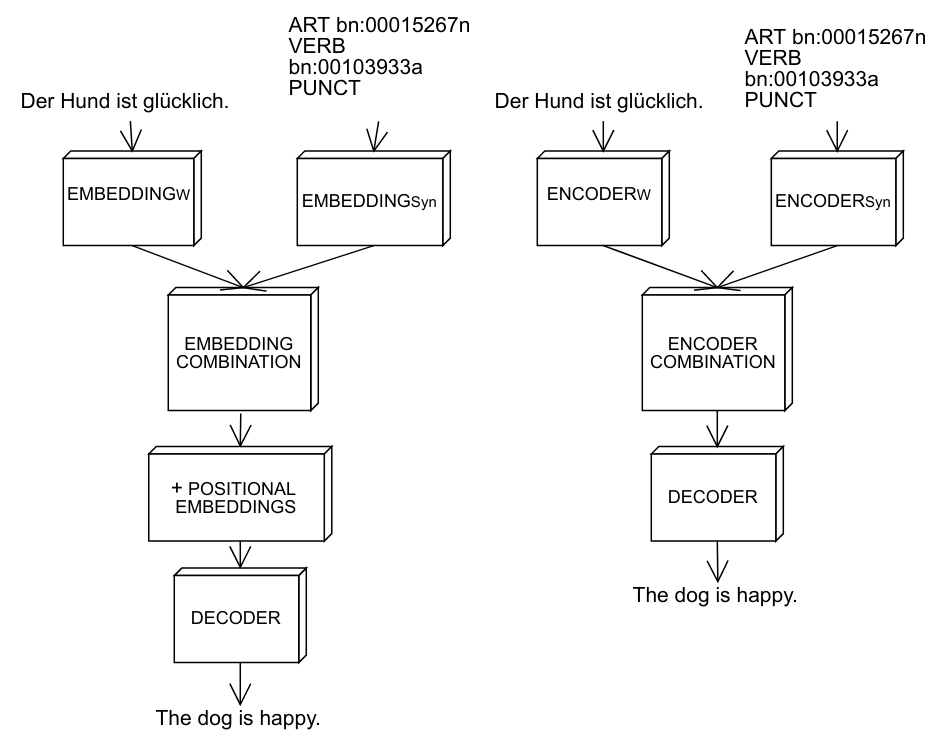}
}{%
  \caption{\label{figure1}1-encoder and N-encoders models.}%
}
\begin{figure}
\begin{floatrow}

\capbtabbox{%
\begin{tabular} {|c|c|c|c|}
\hline
%\multicolumn{4}{|c|}{\textsc{Devtest set}}\\
\multicolumn{4}{|c|}{\textsc{IWSLT14}}\\
  \hline\rule{-2pt}{15pt}
  
  \textsc{Model} & \textsc{Comb.}\texttt{*}& \textsc{Feature} & \textsc{BLEU}\\
  \hline\rule{-4pt}{10pt}
  Baseline & - & - & 34.08\\ 
    Lemmas & - & -  & 29.83 \\
  \hline
  
  1-encoder & Sum & Lemmas & \textbf{34.35} \\
%1-encoder & Concat & PoS& ? \\
  1-encoder & Sum & Babelnet & 33.66 \\ \hline
 %   1-encoder & Concat & Linguistic& ? \\ \hline
  %1-encoder & Sum & Lemmas & \textbf{34.35} \\
  %1-encoder & Sum & BabelNet & 33.66 \\
  1-encoder & Concat & Lemmas & 27.10 \\
  N-encoders & Concat & Lemmas & 33.58 \\
  %N-encoders & Concat & BabelNet & rr \\
  N-encoders & Sum & Lemmas & 9.71 \\ \hline
  %N-encoders & Sum & BabelNet &  rr \\
  % 1-encoder & Sum & BabelNet& 33.66 \\
%   1-encoder & Sum & PoS& ? \\

  \hline
%\end{tabular}
%\begin{verbatim}
%\end{verbatim}
%\begin{tabular} {|c|c|c|c|}
\hline
%\multicolumn{4}{|c|}{\textsc{Test set}}\\
\multicolumn{4}{|c|}{\textsc{IWSLT16}}\\
%  \hline\rule{-2pt}{15pt}
  
%  \textsc{Model} & \textsc{Comb.}\texttt{*}& \textsc{Feature} & \textsc{BLEU}\\
  \hline\rule{-4pt}{10pt}
  Baseline & - & - & 36.67 \\

  1-encoder & Sum & Lemmas & \textbf{37.46}  \\
  \hline\hline
  \multicolumn{4}{|c|}{\textsc{FLoReS}}\\ \hline
   Baseline  & - & - & 3.06\\ \hline
  1-encoder  & Sum & Lemmas & \textbf{4.27}\\ \hline
\end{tabular}
}{%
  \caption{\label{table1}BLEU results. In bold, best results.}%
}

\end{floatrow}
\end{figure}

\begin{comment}
\begin{center}
\begin{figure}[h]
 %\centering
%  \begin{subfigure}{.5\textwidth}
 \centering
  \subfloat{\includegraphics[width=.30\linewidth]{factoredEmbSyn.png}}
  % \caption{A subfigure}
  %\label{fig:sub2}
 %  \end{subfigure}
%\begin{subfigure}{.5\textwidth}
\centering
  \subfloat{ \includegraphics[width=.30\linewidth]{factoredEncoderSyn.png}}
  %\caption{A subfigure}
  %\label{fig:sub2}
%   \end{subfigure}
  \caption{1-encoder and N-encoders.}%: In this case, with German BabelNet tags (or PoS when not available).}
  \label{figure1}
\end{figure}
\end{center}
\end{comment}
%\begin{figure}[h]
  %\centering
  %\includegraphics[width=5cm,clip]{factoredEncoderSyn.png}
  %\caption{Multiple-encoder Factored Transformer}%: In this case, with German BabelNet tags (or PoS when not available).}
  %\label{figure2}
%\end{figure}
\section{Linguistic Features} 
%\label{sec:features}

An arbitrary number of features can be injected into the Factored Transformer, provided they are aligned with words. %In this work, we suggest using linguistic or semantic features, as in the previous works we described above, even though other alternatives could be considered (for instance, domain-specific features for domain adaptation). 
As follows we describe how linguistic features were extracted and how they were aligned at the subword level.

%\subsection{Feature Extraction}

\paragraph{Linguistic tagging with StanfordNLP:} The corpus was tagged with linguistic information, namely lemmas, part-of-speech (PoS), word dependencies and morphological features,  using StanfordNLP \cite{qi2018universal}, and aligned with respect to the original tokenization.

\paragraph{Synsets extraction:} BabelNet's API retrieves all possible \textit{synsets} (semantic identifiers) that a given token may have. Babelfy \cite{Moroetal:14tacl} is a word sense disambiguation service based on BabelNet that retrieves the disambiguated synset for each token depending on the sentence-level context. We split the corpus into chunks such that the daily usage limits of the API were not exceeded and no sentence was split in half (because otherwise Babelfy would have missed the context). Babelfy returns a list of all the detected synsets with their character offsets, and they must be assigned and aligned to the original tokenization of the corpus. The following step was performed to resolve multiword synset conflicts since in the case of synsets composed of more than one token, Babelfy may retrieve one individual synset for each token and a collective one. We decided to prioritize the synset with the largest number of tokens since it seemed to give the most disambiguated information (e.g. the synset \textit{semantic network} gives more specific information than the individual synsets \textit{semantic} and \textit{network}). For the tokens in the corpus that do not have an assigned synset (e.g. articles or punctuation marks), we assign a backup syntactic feature, namely, part-of-speech.%, since many tokens do not have a synset because they are never supposed to (e.g. articles or punctuation marks). %We found that about 70\% of the tokens in the corpus we used did not have any assigned synset, which could be problematic. For this reason, we assigned a backup linguistic feature, namely Part-Of-Speech, since many tokens do not have a synset because they are never supposed to (e.g. articles or punctuation marks).%, and this way the reason why the token did not have any assigned synset would be encoded.

\paragraph{Feature Alignment at the Subword-level:}
%\paragraph{Subword feature alignment and encoding} 
To obtain state-of-the-art results in NMT, subwords (typically, BPE) is usually required. This presents a challenge with regard to word features since they must be aligned with the words themselves. The following alternatives were implemented and experimented with: just repeating the word features for each subword; using the BPE symbol in word features, in the same manner this tag is used in BPE for splitting subwords; and subword tags. This last approach was used in \cite{DBLP:journals/corr/SennrichH16} and it consisted of repeating the word features for each subword and introducing a new factor, subword tags, to encode the position of the subword in the original word. The 4 possible tags are: B (beginning of subword), I (intermediate subword), E (end of the subword) and O (the word was not split). This approach is not compatible with the multiencoder architecture.

\section{Experimental Framework and Results}

%\subsection{Data and Implementation}

\paragraph{Data:} Experiments were conducted with a pair composed of similar languages, the German-to-English translation direction of the IWSLT14 \cite{cettolo2014report}, which is a low-resource dataset (the training set contains about 160,000 sentences). For cleaning and tokenizing, we use the data preparation script proposed by the authors of Fairseq \cite{ott2019fairseq}. We took the test sets from the corpus released for IWSLT14 and IWSLT16. The former was used to test the best configuration, and the latter was used to see the improvement of this configuration in another set. A joint BPE (ie. German and English share subwords) of 32,000 operations is learned from the training data, with a threshold of 50 occurrences for the vocabulary. Other experiments were conducted with the English-to-Nepali translation direction of the FLoRes Low Resource MT Benchmark \cite{DBLP:journals/corr/abs-1902-01382}. Although this pair has more sentences than the previous one (564,000 parallel sentences), it is considered to be extremely low-resource and far more challenging because of the lack of similarity between the involved languages. In this case, we learn a joint BPE of 5000 operations (both with an algorithm based on BPE, sentencepiece \cite{kudo-richardson-2018-sentencepiece}, as proposed by the FLoRes authors, and with the original BPE algorithm). %In the case of sentencepiece, subwords coming from different words can be mapped to the same subword, which makes it unfeasible to align the tags without merging different tags into a single one.

%{\color{red} aqui explicaria els dos datasets posant emfasis en les diferencies un es low-resourced i l'altre es extremely low-resourced, un es de llengues de la mateixa familia linguistica, l'altre no...}

%\paragraph{Parameters and Configurations} {\color{red} aqui hauries de descriure el baseline system, i els parametres d'entrenament i tambe els parametres de les dues arquitectures testejades}

\paragraph{Parameters and Configurations:} In the case of German-to-English, we used the Transformer architecture with the hyperparameters proposed by the Fairseq authors: specifically, 6 layers in the encoder and the decoder, 4 attention heads, embedding sizes of 512 and 1024 for the feedforward expansion size, a dropout of $0.3$ and a total batch size of 4000 tokens, with a label smoothing of $0.1$. For English-to-Nepali, we used the baseline proposed by the FLoRes authors: specifically, 5 layers in the encoder and the decoder, 2 attention heads, embedding sizes of 512 and 2048 for the feedforward expansion size and a total batch size of 4000 tokens, with a label smoothing of 0.2. In both cases, we used the Transformer architecture with the corresponding parameters we described above as the respective baseline systems, and we introduced the modifications of the Factored Transformer without modifying the rest of the architecture and parameters. As mentioned previously, linguistic features were obtained through StanfordNLP \cite{qi2018universal}, except the Babelnet synsets. In the case of the latter, we found that approximately 70\% of the tokens in the corpus we used did not have an assigned synset and were therefore assigned part-of-speech.%, since many tokens do not have a synset because they are never supposed to (e.g. articles or punctuation marks).
%We started by reproducing the baseline, which is very strong even if it is a low-resource dataset, because of the Transformer, the shared BPE and the fact that German and English are similar languages.

\paragraph{Preliminary experiments:} We experimented with %different configurations for the embedding sizes, vocabulary thresholds, 
BPE alignment strategies (including the approaches from section 4.2), and linguistic features extracted from Stanford tagger (lemmas, part-of-speech, word dependencies, morphological features). %We found that large embedding sizes for features with small vocabulary led to overfitting, and that the most promising classical linguistic features were lemmas.
%Once we had done the preliminary research, 
The preliminary experiments showed that %the most important parameter was the feature embedding size, since large embedding sizes for features with a small vocabulary led to overfitting. In contrast,
BPE alignment strategies were not very relevant, so we adopted the alignment with BPE by repeating the word feature. In addition, we found that the most promising linguistic feature was lemmas \cite{DBLP:journals/corr/SennrichH16}.

\paragraph{Reported results:} After the preliminary research, we report experiments with features (lemmas and synsets), architectures (1-encoder and N-encoders systems), and combination strategies (concatenation and summation).
Table \ref{table1} shows the performance of the baseline and the baseline architecture but with lemmas instead of the original words. We report how different features (lemmas or BabelNet) compare for a given architecture. %We tested lemmas as linguistic feature because both in our preliminary experiments and in \cite{DBLP:journals/corr/SennrichH16}
%, which includes emmas, PoS, word dependencies and morphological features) compare. 
%Coherently with  \cite{DBLP:journals/corr/SennrichH16}, 
%the best performing feature is lemmas. BabelNet performs worse than lemmas. 
Then, for the best feature, lemmas, Table \ref{table1} compares different architectures, and it is shown that the best architecture is the 1-encoder with summation. %Then, results show that lemmas outperform BabelNet and linguistic information (lemmas, part-of-speech, word dependencies and morphological features). 
Finally, the best performing system (lemmas with a 1-encoder and summation) is evaluated in another test set, IWSLT16. %(using weight-averaging of the last 10 epochs for both the baseline and our best architecture).
The selected model is relatively efficient, because it only needs an additional embedding layer with respect to the baseline.%, while the total embedding size does not have to be increased because the embeddings are summed instead of concatenated.

Once we had found that the 1-encoder Factored Transformer with summation and lemmas was a solid configuration for low-resource settings, we applied this combination the more challenging Facebook Low Resource (FLoRes) MT Benchmark. Specifically, we wanted to compare how this architecture performs against the baseline reported in the original work of this benchmark. The authors report the results before applying backtranslation and with sentencepiece, which is 4.30 BLEU. We reproduced that baseline and we got slightly better results (up to 4.38 BLEU). However, our system is designed to work with BPE, not sentencepiece, which is more challenging to align to features (since subwords coming from different words can be combined into a single token). Table \ref{table1} shows that our configuration clearly outperformed the baseline with BPE (almost 40\% up), and was very close to the results with sentencepiece.

\paragraph{Discussion:} The 1-encoder system outperforms the N-encoder one. We hypothesize that the N-encoder architecture does not give good results because a completely disentangled representation for each feature is being learned, and this is not an effective strategy for factored NMT. Therefore, it is better to combine features and words at the embedding level, not at the hidden-state level. In the case of N-encoder with concatenation, if the linguistic features are not useful if they come from a different encoder, the decoder at least can learn to ignore them. In the case of the N-encoder architecture with sum, since the outputs from different encoders, which are potentially in very different spaces, are summed, it is tough for the decoder to interpret the vectors. In this case the decoder should learn to undo a sum, which is more difficult than just learning to ignore half of the vector (i.e., assigning low weights). In the case of the 1-encoder architecture, summation gives a much more compact representation. Summing lemmas allows the decoder layers to have a dimension of 512 (instead of doubling that, which may overfit). Regarding the reasons why lemmas outperform synsets, we believe that the problem comes from the fact that  a significant proportion of the tokens do not get a synset. Instead, we can tag all words with lemmas. Besides, the use of synsets (BabelNet) intends to help at disambiguating, but the Transformer is already good at this task \cite{DBLP:journals/corr/abs-1808-08946}.

\section{Conclusions}
%To sum up, we have successfully adapted the current state-of-the-art NMT architecture, the Transformer, to being able to work with factors, that is, to being input features apart from the original source sequences. 
%We have implemented two variants, one with multiple embedding layers and the other one with multiple encoders, and two combination strategies, %namely 
%concatenation and summation. We have extracted semantic information from linked data, tagged the corpus with classical linguistic features and used these as features for factored NMT. %We have tagged the corpus with classical linguistic features as well. 
%We have conducted experiments with different combinations in order to determine the optimal architecture configuration, the most useful features and the best way of aligning and encoding the features with BPE.

We have shown that the Transformer can take advantage of linguistic features but not synsets. %, which is a result that to the best of our knowledge was not published so far. 
We conclude that the best configuration for the Factored Transformer was the 1-encoder model (with multiple embedding layers) with summation instead of concatenation. % and lemmas. Semantic features extracted from linked data did not obtain noticeable improvements.
%At the end, we have consistently shown improvements in two different benchmarks: IWSLT and FLoReS. 
For the German-to-English IWSLT task, the best configuration for the Factored Transformer shows an improvement of 0.8 BLEU, and for the extremely low-resourced English-to-Nepali task, the improvement is 1.2 BLEU. %\footnote{NOTE to reviewers: we are planning to release the code after revisions and bidding period.}.
In future work, we suggest adapting the alignment algorithm to sentencepiece by combining features coming from different words into a single feature, provided their respective subwords have been merged into a single token. In addition, whether the advantage provided by linguistic features still holds once backtranslation has been applied and up to what point this holds should be investigated.

\section*{Acknowledgements}
The authors want to specially thank Mercedes Garc\'ia and Cristina Espa\~{n}a for the insightful discussions.  
This work is supported by the European Research Council (ERC) under the European Union's Horizon 2020 research and innovation programme (grant agreement No. 947657).
\label{main:ref}

\bibliographystyle{acl_natbib}
\bibliography{lrec2020W-xample}

\begin{thebibliography}{19}
\expandafter\ifx\csname natexlab\endcsname\relax\def\natexlab#1{#1}\fi

\bibitem[{Cettolo et~al.(2014)Cettolo, Niehues, St{\"u}ker, Bentivogli, and
  Federico}]{cettolo2014report}
Mauro Cettolo, Jan Niehues, Sebastian St{\"u}ker, Luisa Bentivogli, and
  Marcello Federico. 2014.
\newblock Report on the 11th iwslt evaluation campaign, iwslt 2014.
\newblock In \emph{Proceedings of the International Workshop on Spoken Language
  Translation, Hanoi, Vietnam}, page~57.

\bibitem[{Currey and Heafield(2018)}]{currey-heafield-2018-multi}
Anna Currey and Kenneth Heafield. 2018.
\newblock \href {https://doi.org/10.18653/v1/D18-1327} {Multi-source syntactic
  neural machine translation}.
\newblock In \emph{Proceedings of the 2018 Conference on Empirical Methods in
  Natural Language Processing}, pages 2961--2966, Brussels, Belgium.
  Association for Computational Linguistics.

\bibitem[{Currey and Heafield(2019)}]{currey-heafield-2019-incorporating}
Anna Currey and Kenneth Heafield. 2019.
\newblock \href {https://doi.org/10.18653/v1/W19-5203} {Incorporating source
  syntax into transformer-based neural machine translation}.
\newblock In \emph{Proceedings of the Fourth Conference on Machine Translation
  (Volume 1: Research Papers)}, pages 24--33, Florence, Italy. Association for
  Computational Linguistics.

\bibitem[{Du et~al.(2016)Du, Way, and Zydron}]{du-etal-2016-using}
Jinhua Du, Andy Way, and Andrzej Zydron. 2016.
\newblock \href {https://www.aclweb.org/anthology/L16-1002} {Using {B}abel{N}et
  to improve {OOV} coverage in {SMT}}.
\newblock In \emph{Proceedings of the Tenth International Conference on
  Language Resources and Evaluation ({LREC}'16)}, pages 9--15, Portoro{\v{z}},
  Slovenia. European Language Resources Association (ELRA).

\bibitem[{Espa{\~{n}}a-Bonet and van
  Genabith(2018)}]{espanaVanGenabith:LREC:2018}
Cristina Espa{\~{n}}a-Bonet and Josef van Genabith. 2018.
\newblock Multilingual semantic networks for data-driven interlingua seq2seq
  systems.
\newblock In \emph{Proceedings of the LREC 2018 MLP-MomenT Workshop}, pages
  8--13, Miyazaki, Japan.

\bibitem[{Garc{\'{\i}}a{-}Mart{\'{\i}}nez
  et~al.(2016)Garc{\'{\i}}a{-}Mart{\'{\i}}nez, Barrault, and
  Bougares}]{DBLP:journals/corr/Garcia-Martinez16}
Mercedes Garc{\'{\i}}a{-}Mart{\'{\i}}nez, Lo{\"{\i}}c Barrault, and Fethi
  Bougares. 2016.
\newblock \href {http://arxiv.org/abs/1609.04621} {Factored neural machine
  translation}.
\newblock \emph{CoRR}, abs/1609.04621.

\bibitem[{Guzm{\'{a}}n et~al.(2019)Guzm{\'{a}}n, Chen, Ott, Pino, Lample,
  Koehn, Chaudhary, and Ranzato}]{DBLP:journals/corr/abs-1902-01382}
Francisco Guzm{\'{a}}n, Peng{-}Jen Chen, Myle Ott, Juan Pino, Guillaume Lample,
  Philipp Koehn, Vishrav Chaudhary, and Marc'Aurelio Ranzato. 2019.
\newblock \href {http://arxiv.org/abs/1902.01382} {Two new evaluation datasets
  for low-resource machine translation: Nepali-english and sinhala-english}.
\newblock \emph{CoRR}, abs/1902.01382.

\bibitem[{Koehn and Hoang(2007)}]{koehn-hoang-2007-factored}
Philipp Koehn and Hieu Hoang. 2007.
\newblock \href {https://www.aclweb.org/anthology/D07-1091} {Factored
  translation models}.
\newblock In \emph{Proceedings of the 2007 Joint Conference on Empirical
  Methods in Natural Language Processing and Computational Natural Language
  Learning ({EMNLP}-{C}o{NLL})}, pages 868--876, Prague, Czech Republic.
  Association for Computational Linguistics.

\bibitem[{Kudo and Richardson(2018)}]{kudo-richardson-2018-sentencepiece}
Taku Kudo and John Richardson. 2018.
\newblock \href {https://doi.org/10.18653/v1/D18-2012} {{S}entence{P}iece: A
  simple and language independent subword tokenizer and detokenizer for neural
  text processing}.
\newblock In \emph{Proceedings of the 2018 Conference on Empirical Methods in
  Natural Language Processing: System Demonstrations}, pages 66--71, Brussels,
  Belgium. Association for Computational Linguistics.

\bibitem[{Libovick{\'y} et~al.(2018)Libovick{\'y}, Helcl, and
  Mare{\v{c}}ek}]{libovicky-etal-2018-input}
Jind{\v{r}}ich Libovick{\'y}, Jind{\v{r}}ich Helcl, and David Mare{\v{c}}ek.
  2018.
\newblock \href {https://doi.org/10.18653/v1/W18-6326} {Input combination
  strategies for multi-source transformer decoder}.
\newblock In \emph{Proceedings of the Third Conference on Machine Translation:
  Research Papers}, pages 253--260, Brussels, Belgium. Association for
  Computational Linguistics.

\bibitem[{Moro et~al.(2014)Moro, Raganato, and Navigli}]{Moroetal:14tacl}
Andrea Moro, Alessandro Raganato, and Roberto Navigli. 2014.
\newblock {Entity Linking meets Word Sense Disambiguation: a Unified Approach}.
\newblock \emph{Transactions of the Association for Computational Linguistics
  (TACL)}, 2:231--244.

\bibitem[{Navigli and Ponzetto(2012)}]{NavigliPonzetto:12aij}
Roberto Navigli and Simone~Paolo Ponzetto. 2012.
\newblock {B}abel{N}et: {T}he automatic construction, evaluation and
  application of a wide-coverage multilingual semantic network.
\newblock \emph{Artificial Intelligence}, 193:217--250.

\bibitem[{Ott et~al.(2019)Ott, Edunov, Baevski, Fan, Gross, Ng, Grangier, and
  Auli}]{ott2019fairseq}
Myle Ott, Sergey Edunov, Alexei Baevski, Angela Fan, Sam Gross, Nathan Ng,
  David Grangier, and Michael Auli. 2019.
\newblock fairseq: A fast, extensible toolkit for sequence modeling.
\newblock In \emph{Proceedings of NAACL-HLT 2019: Demonstrations}.

\bibitem[{Qi et~al.(2018)Qi, Dozat, Zhang, and Manning}]{qi2018universal}
Peng Qi, Timothy Dozat, Yuhao Zhang, and Christopher~D. Manning. 2018.
\newblock \href {https://nlp.stanford.edu/pubs/qi2018universal.pdf} {Universal
  dependency parsing from scratch}.
\newblock In \emph{Proceedings of the {CoNLL} 2018 Shared Task: Multilingual
  Parsing from Raw Text to Universal Dependencies}, pages 160--170, Brussels,
  Belgium. Association for Computational Linguistics.

\bibitem[{Sennrich and Haddow(2016)}]{DBLP:journals/corr/SennrichH16}
Rico Sennrich and Barry Haddow. 2016.
\newblock \href {http://arxiv.org/abs/1606.02892} {Linguistic input features
  improve neural machine translation}.
\newblock \emph{CoRR}, abs/1606.02892.

\bibitem[{Sennrich et~al.(2016)Sennrich, Haddow, and
  Birch}]{sennrich-etal-2016-neural}
Rico Sennrich, Barry Haddow, and Alexandra Birch. 2016.
\newblock \href {https://doi.org/10.18653/v1/P16-1162} {Neural machine
  translation of rare words with subword units}.
\newblock In \emph{Proceedings of the 54th Annual Meeting of the Association
  for Computational Linguistics (Volume 1: Long Papers)}, pages 1715--1725,
  Berlin, Germany. Association for Computational Linguistics.

\bibitem[{Tang et~al.(2018)Tang, M{\"{u}}ller, Rios, and
  Sennrich}]{DBLP:journals/corr/abs-1808-08946}
Gongbo Tang, Mathias M{\"{u}}ller, Annette Rios, and Rico Sennrich. 2018.
\newblock \href {http://arxiv.org/abs/1808.08946} {Why self-attention? {A}
  targeted evaluation of neural machine translation architectures}.
\newblock \emph{CoRR}, abs/1808.08946.

\bibitem[{Tebbifakhr et~al.(2018)Tebbifakhr, Agrawal, Negri, and
  Turchi}]{tebbifakhr-etal-2018-multi}
Amirhossein Tebbifakhr, Ruchit Agrawal, Matteo Negri, and Marco Turchi. 2018.
\newblock \href {https://doi.org/10.18653/v1/W18-6471} {Multi-source
  transformer with combined losses for automatic post editing}.
\newblock In \emph{Proceedings of the Third Conference on Machine Translation:
  Shared Task Papers}, pages 846--852, Belgium, Brussels. Association for
  Computational Linguistics.

\bibitem[{Vaswani et~al.(2017)Vaswani, Shazeer, Parmar, Uszkoreit, Jones,
  Gomez, Kaiser, and Polosukhin}]{vaswani2017attention}
Ashish Vaswani, Noam Shazeer, Niki Parmar, Jakob Uszkoreit, Llion Jones,
  Aidan~N Gomez, {\L}ukasz Kaiser, and Illia Polosukhin. 2017.
\newblock \href {https://papers.nips.cc/paper/7181-attention-is-all-you-need}
  {Attention is all you need}.
\newblock In \emph{Advances in neural information processing systems}, pages
  5998--6008.

\end{thebibliography}

%\section{Language Resource References}

%\color{red} Com a language resource references potser hauríem d'incloure BabelNet, Stanford tagger?

%\section{Language Resource References}
%\label{lr:ref}
%\bibliographystylelanguageresource{lrec}
%\bibliographylanguageresource{lrec2020W-xample}

\end{document}